\documentclass[runningheads]{llncs}

\usepackage{eccv}

\usepackage{eccvabbrv}

\usepackage{graphicx}
\usepackage{booktabs}
\usepackage[scr=boondox]{mathalfa}
\usepackage{amssymb}
\usepackage{amsmath,amssymb}
\usepackage{float}
\usepackage{relsize}

\usepackage{pifont}
\usepackage{booktabs}
\usepackage[title]{appendix}
\usepackage{nicematrix}
\usepackage{algorithm}
\usepackage{algpseudocode}
\usepackage{nameref}
\usepackage[dvipsnames]{xcolor}
\usepackage[accsupp]{axessibility}  

\usepackage{hyperref}

\usepackage{orcidlink}

\begin{document}

\title{Instant Uncertainty Calibration of NeRFs Using a Meta-Calibrator} 
\titlerunning{Instant Uncertainty Calibration of NeRFs Using a Meta-Calibrator}

\author{Niki Amini-Naieni\inst{1}\orcidlink{0009-0007-8301-1010} \and
Tomas Jakab\inst{1}\and
Andrea Vedaldi\inst{1}\orcidlink{0000-0003-1374-2858} \and
Ronald Clark\inst{1}}

\authorrunning{N.~Amini-Naieni et al.}

\institute{University of Oxford}

\maketitle
\begin{abstract}
\label{sec:abstract}
Although Neural Radiance Fields (NeRFs) have markedly improved novel view synthesis, accurate uncertainty quantification in their image predictions remains an open problem. The prevailing methods for estimating uncertainty, including the state-of-the-art Density-aware NeRF Ensembles (DANE) \cite{da-nerf}, quantify uncertainty without calibration. This frequently leads to over- or under-confidence in image predictions, which can undermine their real-world applications. In this paper, we propose a method which, for the first time, achieves calibrated uncertainties for NeRFs. To accomplish this, we overcome a significant challenge in adapting existing calibration techniques to NeRFs: a need to hold out ground truth images from the target scene, reducing the number of images left to train the NeRF. This issue is particularly problematic in sparse-view settings, where we can operate with as few as three images. To address this, we introduce the concept of a meta-calibrator that performs uncertainty calibration for NeRFs with a single forward pass without the need for holding out any images from the target scene. Our meta-calibrator is a neural network that takes as input the NeRF images and uncalibrated uncertainty maps and outputs a scene-specific calibration curve that corrects the NeRF’s uncalibrated uncertainties. We show that the meta-calibrator can generalize on unseen scenes and  achieves well-calibrated and state-of-the-art uncertainty for NeRFs, significantly beating DANE and other approaches. This opens opportunities to improve applications that rely on accurate NeRF uncertainty estimates such as next-best view planning and potentially more trustworthy image reconstruction for medical diagnosis. The code is available at \href{https://niki-amini-naieni.github.io/instantcalibration.github.io/}{https://niki-amini-naieni.github.io/instantcalibration.github.io/}.
\end{abstract}
\section{Introduction}
\label{sec:intro}

Recent advancements in scene representations have led to promising new approaches for novel view synthesis and scene reconstruction. Among these, Neural Radiance Fields (NeRFs)~\cite{nerf} have emerged as a particularly powerful tool, offering unprecedented levels of realism and detail in rendered images. The core idea behind NeRFs is to represent a scene as a continuous vector-valued function, parameterized by a neural network, that maps spatial coordinates and view directions to color and density values. This approach enables the creation of detailed 3D representations from sets of 2D images, revolutionizing 3D reconstruction.\\

\begin{figure}[t!]
    \centering
    \includegraphics[width=0.7\linewidth]{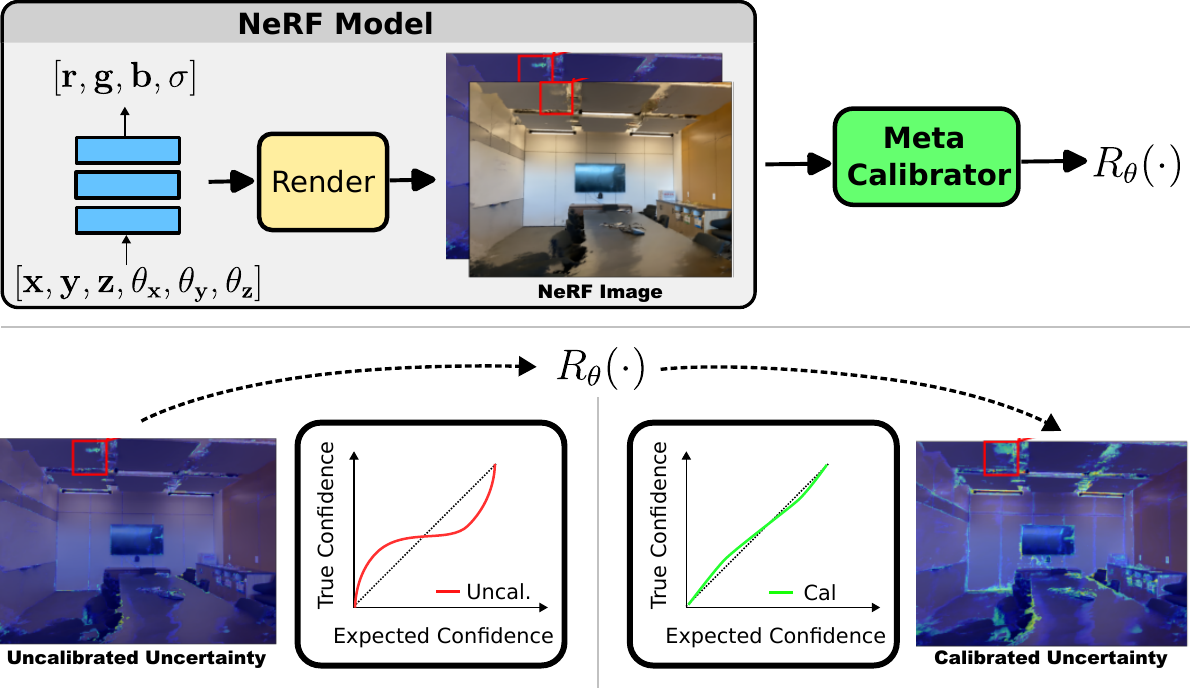}
    \caption{We propose a method for efficiently calibrating the uncertainties from NeRF models. Our approach is based on a meta-calibrator that takes as input features from the rendered NeRF images and uncalibrated uncertainty maps and predicts the calibration function, \( R_{\boldsymbol{\theta}}(\cdot) \), for the NeRF model. Our meta-calibrator generalizes across scenes so it only needs to be trained once, and can predict the calibration function in a single forward pass without any ground truth data from the target scene.}
    \label{fig:overview}
\end{figure}
However, despite their impressive capabilities, traditional NeRF models lack an essential component: an accurate measure of uncertainty in their predictions. Accurate uncertainties are crucial for applying NeRFs to safety-critical problems such as MRI image reconstruction from sparse data \cite{nerf_mri}, where unreliable confidence estimates could lead to misdiagnosis. More accurate uncertainties could also enhance practical methods such as uncertainty-guided next-best view planning techniques\cite{nextbestview}. While prior approaches have attempted to estimate NeRF uncertainties \cite{cf-nerf, s-nerf, bayesrays, wildnerf, da-nerf}, they all overlook the problem of calibration. Thus, the uncertainties they output are not as accurate as they could be.

In particular, the state-of-the-art uncertainty estimation method for NeRFs, Density-aware NeRF Ensembles (DANE) \cite{da-nerf}, produces uncalibrated uncertainties. As a result, the confidence intervals and variances do not match the true confidences, meaning it has limited applicability to real-world problems. This constraint is significant as it restricts the use of NeRFs in safety-critical and sparse-data settings, where knowing the confidence in predictions is crucial.\\

The best NeRF methods for the sparse-view setting overlook the problem of calibration as well. FlipNeRF \cite{flipnerf}, the state-of-the-art technique for sparse-view reconstruction, uses uncertainties from an uncalibrated mixture of Laplacians to enhance its training process. Therefore, the uncertainties it outputs at inference as an artifact of training are not accurate. \\

In this paper, we present a novel approach for obtaining calibrated uncertainties for NeRF models in the sparse-view setting. Our strategy integrates the Laplacian mixture from FlipNeRF~\cite{flipnerf} with the calibration techniques by Kuleshov et al~\cite{CalRegr}. However, naively applying the calibration method by Kuleshov et al to FlipNeRF does not work due to a significant challenge of the sparse-view setting: there is a lack of held-out data from the target scene for fitting the calibrator. \textbf{Specifically, holding out just one image for calibration could decrease the size of the training set by over 30 \%, resulting in significant performance degradation of the NeRF. }To overcome this, we make use of a unique observation: while calibration curves exhibit significant variation across scenes, they also demonstrate a significant regularity in their structure. Utilizing this insight, we propose the concept of a meta-calibrator that learns a low-dimensional representation of the NeRF calibration curves and infers them from scene features. We motivate and show why this meta-calibrator is necessary and demonstrate that it achieves more accurate uncertainties than DANE \cite{da-nerf} without holding out any images from the target scene.\\

Specifically, our contributions are: (1) the first investigation into obtaining calibrated uncertainties from NeRFs, (2) a novel meta-calibrator for fitting the calibration model without using held-out data, and (3) experiments on the real-world Local Light Field Fusion (LLFF) \cite{llff} and DTU  \cite{dtu} datasets showing that our meta-calibrator achieves state-of-the-art and well-calibrated uncertainties for real scenes. We also demonstrate that our uncertainties can be leveraged for effective next-best view planning.

\section{Related Work}
\label{sec:rel_work}

\paragraph{\textbf{Neural Radiance Fields (NeRFs).}}
NeRFs \cite{nerf} are a popular method for novel view synthesis. From a set of 2D images, NeRFs learn a neural network representation of a single scene. A trained NeRF model outputs estimates of the volume density and emitted radiance at any 3D location and viewing direction. Novel views can be generated by applying volume rendering \cite{volrend} to the density and radiance values predicted by the NeRF model for points along rays cast into the scene. Due to their simplicity and impressive performance, NeRFs have become a popular technique for solving a variety of rendering problems.\\

Over the last few years, several extensions of NeRFs have been explored \cite{nerfplusplus, animanerf, donerf, nerfvae, nerfperact}. These include speeding up training and inference \cite{f2nerf, sparsevoxelnerf, derf}, modeling dynamic scenes \cite{dynanerf}, learning from a sparse set of training views \cite{mixnerf, flipnerf, mipnerf, dietnerf, regnerf}, and estimating the uncertainty in NeRF predictions \cite{cf-nerf, s-nerf, bayesrays, wildnerf, neurar, da-nerf}. Sparse NeRF methods aim to accurately render novel views when only a few training views are available from the target scene. NeRF uncertainty estimation techniques strive to accurately predict the confidence in the views rendered. Although uncertainty estimation is particularly important in the sparse-view setting, where NeRF renderings are especially unreliable, the main aim of sparse NeRF methods is not to output accurate uncertainties.

\paragraph{\textbf{Sparse-view NeRF Methods.}}
Despite this, recent sparse NeRF methods do produce uncertainties as an artifact of their training process. Both MixNeRF \cite{mixnerf} and FlipNeRF \cite{flipnerf}, the state-of-the-art approach, model the RGB color channels given a ray as independent random variables that follow a mixture model. FlipNeRF further uses the uncertainties of the pixel colors to regularize the training process, producing more accurate image reconstructions at inference. However, neither MixNeRF nor FlipNeRF outputs calibrated uncertainties.\\

Our method significantly extends sparse NeRF methods to obtain more accurate and well-calibrated uncertainties at inference. To benefit from the superior performance of FlipNeRF \cite{flipnerf} at sparse novel view synthesis, we apply the proposed meta-calibrator to the learned distribution from FlipNeRF, producing more accurate uncertainties without sacrificing state-of-the-art image quality. However, our approach can be applied to any NeRF method that outputs uncertainties, so it is distinct from FlipNeRF and MixNeRF. In essence, the proposed meta-calibrator augments sparse NeRF methods to achieve state-of-the-art uncertainty, beating both FlipNeRF and techniques designed explicitly for NeRF uncertainty estimation.

\paragraph{\textbf{Uncertainty in NeRFs.}}
The growing line of methods specifically designed for accurately estimating NeRF uncertainties \cite{cf-nerf, s-nerf, bayesrays, wildnerf, neurar, da-nerf} do not address the problem of calibration. The current state-of-the-art uncertainty estimation technique for NeRFs is Density-aware NeRF Ensembles (DANE) \cite{da-nerf}. DANE adds an epistemic uncertainty term to a naive ensembles approach with five ensemble members. Thus, DANE is very costly as it requires training five NeRFs to obtain uncertainty estimates. Another NeRF uncertainty estimation approach is Stochastic Neural Radiance Fields (S-NeRF) \cite{s-nerf}, which learns a probability distribution over all possible radiance fields by modeling the volume density and radiance as random variables that follow a joint distribution. S-NeRF employs variational inference to sample from an approximation to this distribution and uses the variances of the sampled pixel colors as the estimated uncertainties. Conditional-flow NeRF (CF-NeRF) \cite{cf-nerf} builds on S-NeRF by combining latent variable modeling and conditional normalizing flows to relax the strong constraints S-NeRF imposes over the radiance distribution. Despite the growing number of techniques in this area of study, all prior work does not consider calibration, outputting unreliable uncertainties as a result. Our work achieves more accurate uncertainties than these prior methods by filling the gap of uncertainty calibration for NeRFs and drawing on well-established techniques in calibrated regression \cite{CalRegr}. 

\paragraph{\textbf{Uncertainty Calibration in Deep Learning.}}
While uncertainty calibration has been studied for Bayesian deep learning methods \cite{CalRegr, caluncdeep}, it has not been adapted for or applied successfully to NeRF uncertainty estimates. This may be because NeRFs introduce additional complexity in the uncertainty estimation process as the neural network model needs to be trained \textit{per-scene}. 

This makes uncertainty calibration challenging as methods for calibrated regression\cite{CalRegr} require either using the training set or held-out data to achieve calibrated uncertainties. Using the training set to fit the calibrator results in severe overfitting (see \cref{sec:train_curve_overfit}). Holding out data from the target scene for calibration means there is less data to train the NeRF, making it more inaccurate at novel view synthesis (see \cref{sec:hold_out_data_poor}). Thus, a trivial application of \cite{CalRegr} to NeRFs would not be satisfactory. In this work, we propose the concept of a meta-calibrator that, in contrast to \cite{CalRegr}, does not require held-out data and achieves calibrated uncertainty estimates for NeRFs.\\



\section{Method}
\label{sec:method}


In this paper, we present a method that calibrates NeRF uncertainties. To this end, we propose a novel meta-calibrator that accepts uncalibrated NeRF uncertainties and predicted images as inputs and outputs a scene-specific calibration curve, correcting the uncalibrated confidence levels. An overview of our method can be seen in Fig. \ref{fig:overview}. Crucially, our approach does not require holding out any images from the target scene. Thus, it can be applied to sparse-view settings, where holding out a single image could significantly harm the NeRF's performance. In Section \ref{sec:prelim}, we explain the necessary background concepts, in Section \ref{sec:calc_uncert}, we describe how we obtain the corrected uncertainty values, and in Section \ref{sec:metacal} we detail our meta-calibrator.

\subsection{Preliminaries}\label{sec:prelim}
\paragraph{\textbf{Neural Radiance Fields (NeRFs).}}\label{nerf}
Neural Radiance Fields (NeRFs) \cite{nerf} represent a scene as a continuous vector-valued function with inputs a Cartesian point $\mathbf{x} = (x, y, z)$ and unit viewing direction vector $\mathbf{d} = (u, v, w)$ and outputs an emitted radiance $\boldsymbol{\mathscr{c}} = (\mathscr{r}, \mathscr{g}, \mathscr{b})$ and volume density $\sigma$. By optimizing the weights $\mathbf{\Theta}$ of a neural network approximation $\mathbf{F_{\Theta}}$ to this representation, NeRFs can render the color of any pixel in a synthetic image of the scene. To achieve this, principles from classical volume rendering \cite{volrend} are applied to the radiance and density values estimated by $\mathbf{F_{\Theta}}$ for points along a ray cast from the origin $\mathbf{x_{0}}$ of the virtual camera, through the pixel, and into the scene. More specifically, the expected color $\mathbf{c}(\mathbf{r})$ of a camera ray $\mathbf{r}(t) = \mathbf{x_{0}} + t\mathbf{d}$ with near and far bounds $t_{n}$ and $t_{f}$ is:
\begin{equation}\label{nerf_integral}
\mathbf{c}(\mathbf{r}) = \int_{t_{n}}^{t_{f}}T(t)\sigma(\mathbf{r}(t))\boldsymbol{\mathscr{c}}(\mathbf{r}(t), \mathbf{d})dt \text{,}
\end{equation}

\noindent where $T(t) = e^{-\int_{t_{n}}^{t}\sigma(\mathbf{r}(s))ds}$. The integral in \cref{nerf_integral} is estimated with numerical quadrature to obtain the colors of the pixels in the synthetic image from the NeRF model outputs. Since the numerical quadrature is differentiable, NeRF optimizes $\mathbf{\Theta}$ according to \cref{nerf_integral} with gradient descent. While they perform well at novel view synthesis, conventional NeRFs do not provide an uncertainty associated with their predictions, so extensions like DANE \cite{da-nerf} have been developed.

\paragraph{\textbf{Base NeRF Uncertainties.}}\label{base_uncert}
To obtain the initial uncertainties, we have two options for our base model: (1) DANE \cite{da-nerf}, the state-of-the-art method for NeRF uncertainty estimation or (2) FlipNeRF \cite{flipnerf}, the state-of-the-art method for sparse novel view synthesis. Because DANE requires training five NeRFs per scene and provides poor image quality in the sparse-view setting, we choose FlipNeRF. We show in the experiments that applying our meta-calibrator to FlipNeRF results in more accurate and well-calibrated uncertainties than those output by DANE. Our base FlipNeRF uncertainties are inferred from a mixture of Laplacians with location and scale parameters learned during training.\\

FlipNeRF \cite{flipnerf} models the joint distribution of the color $\mathbf{C} = (R, G, B)$ given a ray $\mathbf{r}$ with a mixture of Laplacians:

\begin{equation}
\label{eq:pred_posterior}
p(\mathbf{C} = \mathbf{c}| \mathbf{r}) = \sum_{j = 1}^{M}\pi_{j}\mathcal{F}(\mathbf{C} = \mathbf{c}; \boldsymbol{\mu_{j}}, \boldsymbol{\beta_{j}}) \text{,}
\end{equation}

\noindent where M is the number of sampled points along the ray $\mathbf{r}$. $\mathcal{F}(\mathbf{C} = \mathbf{c}; \boldsymbol{\mu_{j}}, \boldsymbol{\beta_{j}})$ is the 3D Laplacian probability density with location parameter $\boldsymbol{\mu_{j}} = (\mu_{j}^{R}, \mu_{j}^{G}, \mu_{j}^{B})$ and scale parameter $\boldsymbol{\beta_{i}} = (\beta_{j}^{R}, \beta_{j}^{G}, \beta_{j}^{B})$ evaluated at the color $\mathbf{c}$. More specifically, the mixing coefficients $\{\pi_{j}\}_{j = 1}^{M}$ are the normalized coefficients of the radiance values along a ray in \cref{nerf_integral}, the location parameters $\{\boldsymbol{\mu_{j}}\}_{j = 1}^{M}$ are the estimated RGB radiance values, and the scale parameters $\{\boldsymbol{\beta_{j}}\}_{j = 1}^{M}$ are an additional output of the model. These parameters are optimized by FlipNeRF during training.\\

We now go into detail about how the location and scale parameters are obtained. FlipNeRF \cite{flipnerf} learns these parameters by minimizing the negative log-likelihood of the training set $\mathcal{D} = \{(\mathbf{r_{i}}, \mathbf{c_{i}})\}_{i = 1}^{N}$ containing the rays $\mathbf{r_{i}}$ and colors $\mathbf{c_{i}}$ from the pixels in the ground truth images of the scene assuming the distribution in \cref{eq:pred_posterior}. FlipNeRF additionally minimizes the average scale parameters through an auxiliary uncertainty-aware emptiness loss for reducing floating artifacts. The negative log-likelihood loss and the uncertainty-aware emptiness loss are added to FlipNeRF's total training loss, which incorporates other terms such as the mean squared error. As a result of this training process, the location and scale parameters can be inferred by FlipNeRF at new poses. From these location and scale parameters, we obtain the uncalibrated confidence levels of predicted ray colors.\\

Using the distribution in \cref{eq:pred_posterior}, we can easily compute the confidence level for a given ray $\mathbf{r_{t}}$, which we denote as $F_t$. The CDF for the Laplacian mixture has a closed form expression parameterized by the location and scales output by the pretrained FlipNeRF \cite{flipnerf}. Thus, we can use this CDF to predict the confidence level of the ground truth color $\mathbf{c_{t}}$ of any ray by evaluating it at the given color value, $p_t = F_t(\mathbf{c_t})$. However, these initial confidence levels are uncalibrated and, hence, inaccurate.\\

\paragraph{\textbf{Calibrated regression.}} In \cite{CalRegr}, Kuleshov et al extend calibration methods for classification to regression. They define a forecaster $H: \mathcal{X} \rightarrow (\mathcal{Y} \rightarrow [0, 1])$ as a function that outputs for each $x_{t} \in \mathcal{X}$, a CDF $F_{t}$. Given a pretrained forecaster $H$, they suggest training an auxiliary model $R: [0, 1] \rightarrow[0, 1]$ by fitting $R$ to a recalibration dataset $ D = \left\{\left([H(x_{t})](y_{t}), \hat{P}([H(x_{t})](y_{t}))\right)\right\}_{t = 1}^{T}$, where 
\[\hat{P}(p) = |\{y_{t} : [H(x_{t})](y_{t}) \leq p \text{ for } t = 1, \dotsc, T\}| / T  \] is the empirical confidence level corresponding to the predicted confidence level $p$. The fitted $R$ forms the calibration curve that corrects the expected confidence levels. We can now obtain predictive posterior values that closely match the true confidences using $\hat{F}_t \equiv R \circ F_{t}$ for the test data.\\

However, as mentioned before, there are numerous challenges associated with applying this recalibration procedure. Firstly, note that it requires ground-truth values ($y_t$) for the predictions we are recalibrating. This means that in order to prevent overfitting we need to reserve part of the training dataset specifically for fitting the calibrator (which leaves less data for training the NeRF --- a significant issue if we only have a few input views). Secondly, it does not actually prescribe how to compute a suitable uncertainty value from the predicted distribution. In Section \ref{sec:metacal}, we address the former issue, and in Section \ref{sec:calc_uncert}, we address the latter.\\

\subsection{Calculating uncertainty}\label{sec:calc_uncert}

While the predictive posterior in \cref{eq:pred_posterior} provides a distribution over likely ray colors for a NeRF model, it does not inherently offer a straightforward metric for quantifying uncertainty at a specific point in the reconstruction. Intuitively, as the variance of this distribution increases, so does the uncertainty of the model's output at that point. Therefore, the variance or standard deviation is a popular choice for quantifying the uncertainty \cite{s-nerf, cf-nerf, flipnerf, da-nerf}. However, in the case of the corrected mixture distribution obtained from raymarching, this can be slow to compute especially if it has to be done for each pixel. Therefore, we turn to a metric that can be calculated directly from the calibrated CDF $\hat{F}^t$.\\

We propose to use the interquartile range of each calibrated distribution:
\begin{equation}\label{uncert}
    \kappa^C(\mathbf{r_{t}}) = [\hat{F}^C_t]^{-1}\left(\frac{3}{4}\right) - [\hat{F}^C_t]^{-1}\left(\frac{1}{4}\right)\text{,} 
\end{equation}
where \(\kappa^C(\mathbf{r_{t}}\text{)}\) represents the uncertainty at a ray \(\mathbf{r_{t}}\) in the color channel $C$. This difference provides a measure of the statistical dispersion and thus serves as a robust measure of the spread of the output channel. By averaging the interquartile range over the color channels, we obtain a single scalar value that effectively quantifies the uncertainty of the NeRF model for the given ray. As shown in our experiments, this method is very computational efficient, and it provides an accurate estimate of uncertainty.


\begin{figure}[tb]
  \centering
  \begin{subfigure}{0.45\linewidth}
    \includegraphics[width=\linewidth]{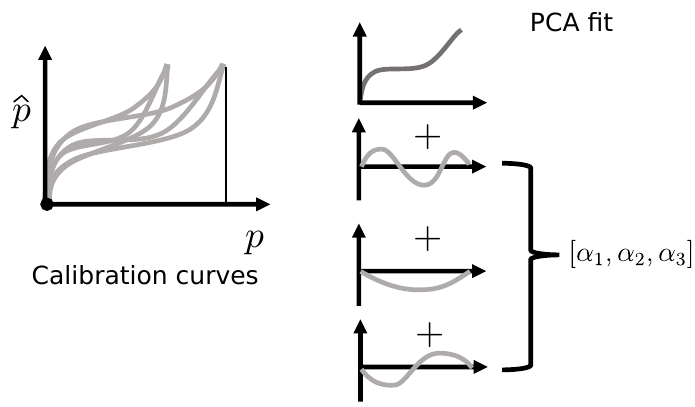}
    \caption{Creating a parametric model of the calibration curves.}
    \label{fig:short-a}
  \end{subfigure}
  \hspace{0.25cm}
  \begin{subfigure}{0.45\linewidth}
    \includegraphics[width=\linewidth]{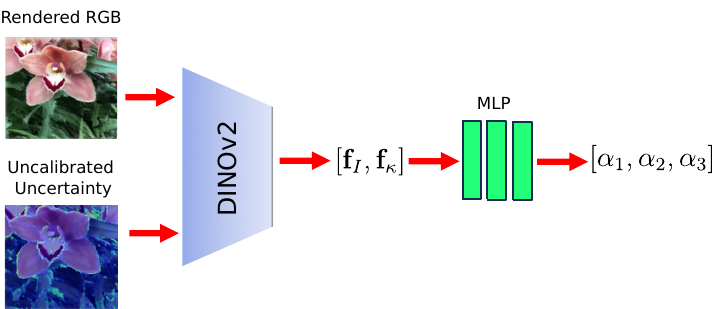}
    \caption{Predicting the curve parameters.}
    \label{fig:short-b}
  \end{subfigure}
  \caption{\textbf{Meta-calibrator design.} In stage (a) we fit a low-dimensional parameteric model of the calibration curves. The meta-calibrator then predicts these curve parameters from rendered images of the scene and their associated uncalibrated uncertainty maps (b).}
  \label{fig:meta-cal}
\end{figure}

\subsection{Meta-calibrator}\label{sec:metacal}
To overcome the challenge that, especially in the sparse-view setting, there is no held-out data available for fitting the calibrator, we propose a novel meta-calibrator that infers the calibration curves from uncalibrated NeRF predictions. To do this, we leverage the insight that the calibration curves demonstrate significant regularity. We posit a low-dimensional model of the calibration curves can be learned and predicted using the images and uncalibrated uncertainty maps inferred by the NeRF, enabling us to estimate the calibration function \textit{without evaluating the empirical confidence levels using held-out data from the target scene}. We now describe this meta-calibrator (illustrated in \cref{fig:meta-cal}) in detail.

\paragraph{\textbf{A Parametric Model for Calibration Curves.}} 
We first fit a low-dimensional representation of the calibration curves using Principal Component Analysis (PCA). To create the training set for learning this representation, we sample held-out images from $K$ scenes and apply the calibration procedure by Kuleshov et al.~\cite{CalRegr} to form $K$ ground truth calibration curves. To construct the training vector $\mathbf{v_k} \in \mathbb{R}^{1 \times M}$ for scene $k \leq K$, we sample $M$ evenly spaced points along its ground truth calibration curve. We find that fitting the PCA model using only a few scenes (21 in our case) provides a good enough approximation to capture the variation in the test curves (see \cref{sec:metacal_design}). Here, $\mathbf{V} = [\mathbf{v}_k] \in \mathbb{R}^{K \times M}$ contains the ground truth calibration curves for the training scenes, with \(K\) representing the number of curves and \(M\) the sample count along each curve.
PCA is then used to determine the basis vectors \(\mathbf{U} = (\mathbf{u_1}, \mathbf{u_2}, ..., \mathbf{u_n})\) and coefficients \(\boldsymbol{\theta} = (\alpha_1, \alpha_2, ..., \alpha_n)\), so that each calibration curve can be represented as:
$ \mathbf{v}_k = \sum_{i=1}^{n} \alpha_i \mathbf{u}_i$. The parameters \(\boldsymbol{\theta}\) fully describe the calibration functions. \\

To find the optimal number of components, we compute the explained variance, and find that in our case most of the variance is explained using only $n=3$ components (see experiments). To ensure the calibrator is monotonically increasing, we derive the final calibration function \( R_{\boldsymbol{\theta}}(\cdot) \) using isotonic regression applied to the curve approximated by $\boldsymbol{\theta} \cdot \mathbf{U}$. The idea is that the low-dimensional representation of the calibration curves encoded in \(\mathbf{U}\) will generalise to new target scenes without any additional scene-specific data.

\paragraph{\textbf{Predicting Calibration Parameters.}}
What remains is to estimate the calibration parameters, \(\boldsymbol{\theta}\), for a new scene. As we do not want to use additional held-out data from the target scene, we propose predicting these parameters using scene-specific features computed from the pretrained NeRF outputs. This approach is motivated by the human ability to visually identify inaccuracies in the renderings such as floaters and unnatural artifacts. Specifically, we use a Multi-Layer Perceptron (MLP) with three layers of output size: [128, 128, 3] and Leaky ReLU activations throughout except the last layer as the meta-calibrator to estimate \(\boldsymbol{\theta}\) given features extracted by the DINOv2 model \cite{oquab2023dinov2} from rendered images (\(\mathbf{f}_I\)) and uncalibrated uncertainty maps (\(\mathbf{f}_\kappa\)). The goal here is to have DINOv2 extract features that describe the rendering imperfections, correlating with the calibration curve. We find that training the MLP model on only a few scenes (30 in our case) allows it to generalize well to new test scenes. Once trained, the meta-calibrator can predict the calibration curve of a new target scene as: $\boldsymbol{\theta} = MLP([\mathbf{f}_I, \mathbf{f}_\kappa])$, without using any additional ground truth data.\\

In summary, the meta-calibrator can correct the confidence levels of the model without requiring ground truth data at any stage, suggesting potential enhancements to applications that rely on uncertainty such as next-best view selection (see Sec. \ref{sec:nbv_exp}).

\section{Experiments}

\label{sec:experiments}
The objective of the experiments is to: 1) validate that our approach achieves more accurate uncertainties (lower negative log-likelihood and calibration error) than state-of-the-art approaches for NeRF uncertainty estimation (\cref{sec:sota_compare}); 2) demonstrate the meta-calibrator improves the accuracy of the uncalibrated uncertainties (decreases both the negative log-likelihood and calibration error) without requiring any held-out data from the target scene (\cref{sec:cal_compare}); 3) explain the motivation for certain meta-calibrator design decisions; and 4) show that our uncertainties can be leveraged for applications such as next-best view planning (\cref{sec:nbv_exp}). 

For additional results showing: 1) why the PCA representation of the calibration curves is necessary; 2) that using the training set results in severe overfitting; 3) that holding out data results in poor performance at image reconstruction; 4) the influence of the number of samples along the ray on the uncertainty quality; and 5) the efficiency of our uncertainty metric (\cref{uncert}) over other approaches, please refer to Appendices \ref{sec:pca_nec}, \ref{sec:train_curve_overfit}, \ref{sec:hold_out_data_poor}, \ref{sec:num_samples}, and \ref{sec:efficiency} respectively.

\paragraph{\textbf{Metrics and calibration curves.}}
We use a variant of the calibration error from \cite{CalRegr} to evaluate the effectiveness of the meta-calibrator. Specifically, given a test set $\mathcal{D} = \{(\mathbf{r_{t}}, \mathbf{c_{t}})\}_{t = 1}^{T} = \{(\mathbf{r_{t}}, (r_{t}, g_{t}, b_{t}))\}_{t = 1}^{T}$, we report:

\begin{equation}\label{cal_err}
ERR = \frac{1}{T} \sum_{t = 1}^{T}(p_{t} - \hat{P}(p_{t}))^{2}\text{, }
\end{equation}

\noindent where $p_{t}$ is the expected confidence level for data point $(\mathbf{r_{t}}, c_{t})$, and $\hat{P}(p_{t})$ is the empirical frequency of data points within that confidence level. More specifically, for each $t \in \{1, \dotsc, T\}$, we set $p_{t} = M_{t}^{C}(c_{t})$ and,
\begin{equation}\label{uncal_cal_err}
\hat{P}(p) = |\{c_{t}: M_{t}^{C}(c_{t}) \leq p\text{ for } t = 1, \dotsc, T\}|/T\text{,}
\end{equation}
where $M \equiv F$ for uncalibrated errors and $M \equiv \hat{F}$ for calibrated errors, and
$C \in \{R, G, B\}$. Note that this formulation of the calibration error is equivalent to the one in \cite{CalRegr} with a confidence level for every unique $p_{t}$ and weights that more significantly penalize errors from frequently predicted confidence levels.

Following \cite{CalRegr}, we plot $\{(p_{t}, \hat{P}(p_{t}))\}_{t = 1}^{T}$ before and after calibration for each color channel to generate calibration curves. A perfectly calibrated forecaster would produce the straight line $p_{t} = \hat{P}(p_{t})$ as each expected confidence level would equal the empirical one. Intuitively, our version of the calibration error is the mean squared vertical distance of points on the calibration curve from a perfectly straight line. If an expected confidence level occurs $N$ times in the test data, its distance is counted $N$ times in the mean.
Following \cite{da-nerf}, we additionally report the negative log-likelihood (NLL) of the test data averaged across all scenes. Following \cite{flipnerf}, we include PSNR and LPIPS \cite{lpips} to evaluate image quality. 

\paragraph{\textbf{Datasets.}} We use 30 scenes from 3 datasets: Realistic Synthetic 360$^{\circ}$ \cite{nerf}, the subset of scenes in DTU \cite{dtu} used in \cite{flipnerf}, and LLFF \cite{llff} for training the meta-calibrator and test it on a hold-out scene from either LLFF or DTU to show it generalizes to new target scenes.

\paragraph{\textbf{Baselines}} We compare our approach against the state-of-the-art method for NeRF uncertainty estimation DANE \cite{da-nerf} as well as other methods in \cref{sec:sota_compare}. In \cref{table:sota}, following \cite{da-nerf}, we implement the naive ensembles approach and DANE using a public implementation of Instant-NGP \cite{bhalgat2022hashnerfpytorch, instant_ngp} and 5 ensemble members. In \cref{sec:cal_compare}, we compare the uncalibrated uncertainties to the uncertainties calibrated by our meta-calibrator. 

\begin{figure}
    \centering
    \begin{subfigure}{0.3\linewidth}
      \includegraphics[width=\linewidth]{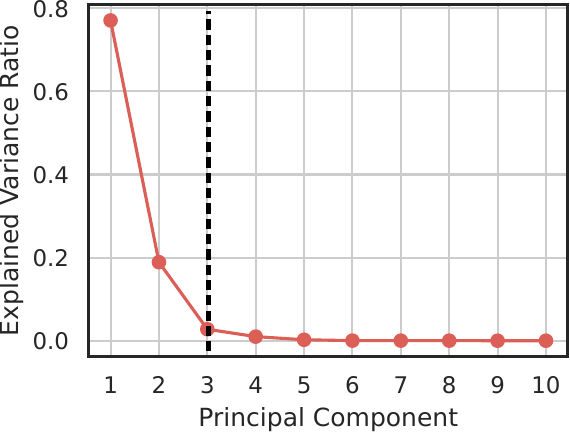}
      \caption{Explained variance plot showing most of the variance of the calibration curves is explained using only 3 PCA components.}
      \label{fig:explained_var}
    \end{subfigure}
    \hfill
    \begin{subfigure}{0.3\linewidth}
      \includegraphics[width=\linewidth]{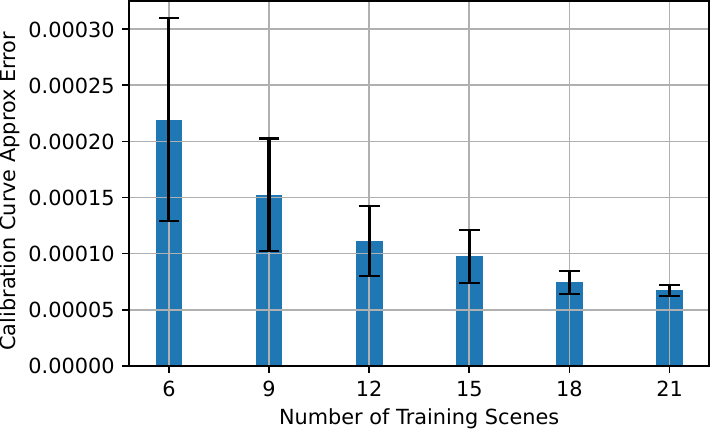}
      \caption{Graph showing calibration error on new scenes is sufficiently low when using 21 scenes to fit 3-coefficient PCA model (i.e. it’s an order of magnitude lower than we can expect from the final calibration, meaning good generalization to unseen scenes).}
      \label{fig:pca_fit_num_scenes}
    \end{subfigure}
    \hfill
    \begin{subfigure}{0.3\linewidth}
      \includegraphics[width=\linewidth]{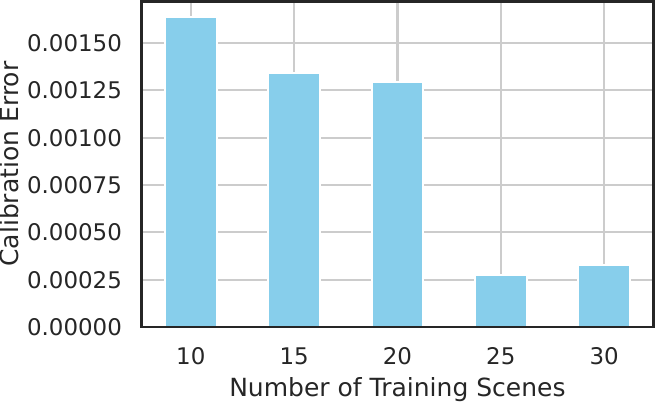}
      \caption{Graph showing increasing number of training scenes for meta-calibrator decreases calibration error for the test scene \emph{T-Rex} from LLFF \cite{llff}, with 30 scenes resulting in good generalization to the held-out scene and well-calibrated uncertainties.}
      \label{fig:meta_cal_fit}
    \end{subfigure}
    \caption{\textbf{Meta-calibrator design decisions.} Results showing using 3 components for Principal Component Analysis (PCA) model of calibration curves, 21 scenes to fit PCA model, and 30 scenes to train meta-calibrator achieves good generalization to new test scenes.}
    \label{fig:meta-cal-design}
\end{figure}

\paragraph{\textbf{Meta-calibrator Design}}
\label{sec:metacal_design}
The results guiding our decisions to use 3 Principal Component Analysis (PCA) components to represent the calibration curves, fit the PCA components using 21 training scenes, and train the meta-calibrator on 30 scenes to predict the PCA coefficients are shown in \cref{fig:meta-cal-design}.

\begin{table}
\begin{center}
\scriptsize
\caption{\label{table:sota}\textbf{Quantitative results on standard sparse NeRF benchmark.} Our proposed approach results in significantly better uncertainties and image quality than the state-of-the-art NeRF uncertainty estimation method DANE \cite{da-nerf} does on the challenging 3-view LLFF \cite{llff} dataset. Specifically, our meta-calibrator reduces the calibration error to 6\% of DANE's calibration error and the negative log-likelihood to be over 100 \% lower than DANE's. Note: \textbf{\emph{lower calibration error (Cal. Err.) and negative log-likelihood (NLL) values indicate more accurate uncertainties.}} Results are averaged over all 8 scenes in LLFF.}
\begin{NiceTabular}{c|c|c|c|c} 
 \hline
 & \multicolumn{2}{c}{Uncertainty} & \multicolumn{2}{c}{Image Quality}\\
 & Cal. Err. & NLL & PSNR & LPIPS\\ 
 & ($\downarrow$) & ($\downarrow$) & ($\uparrow$)& ($\downarrow$)\\ 
  \hline
 Naïve Ens. & 0.0505 & 4.39 & 15.19 & 0.646\\
 DANE \cite{da-nerf} & 0.0441 & 3.75 & 15.19 & 0.646\\
 \textbf{Ours} & \textbf{0.0026} & \textbf{-0.68} & \textbf{19.34} & \textbf{0.235}\\
 \hline
\end{NiceTabular}
\end{center}
\end{table}

\subsection{Comparison to State-of-the-art}
\label{sec:sota_compare}
In this section, we compare our approach to prior methods for NeRF uncertainty estimation. We achieve more accurate uncertainties (94 \% reduction in calibration error and over 100 \% reduction in negative log-likelihood) than those estimated by DANE \cite{da-nerf}, the state-of-the-art method. These results are shown in \cref{table:sota} for the challenging 3-view LLFF \cite{llff} dataset from prior work on sparse novel view synthesis \cite{flipnerf, mixnerf, regnerf} and \cref{table:uncert_sota} for the less challenging version of LLFF from prior work on NeRF uncertainty estimation \cite{cf-nerf, s-nerf, da-nerf, wildnerf}. In \cref{fig:dane_curve_compare}, we compare the calibration curves from our approach to DANE's, illustrating that the meta-calibrator predicts expected confidences that match the true ones while DANE does not.
\begin{figure*}
    \centering
    \includegraphics[width=\linewidth]{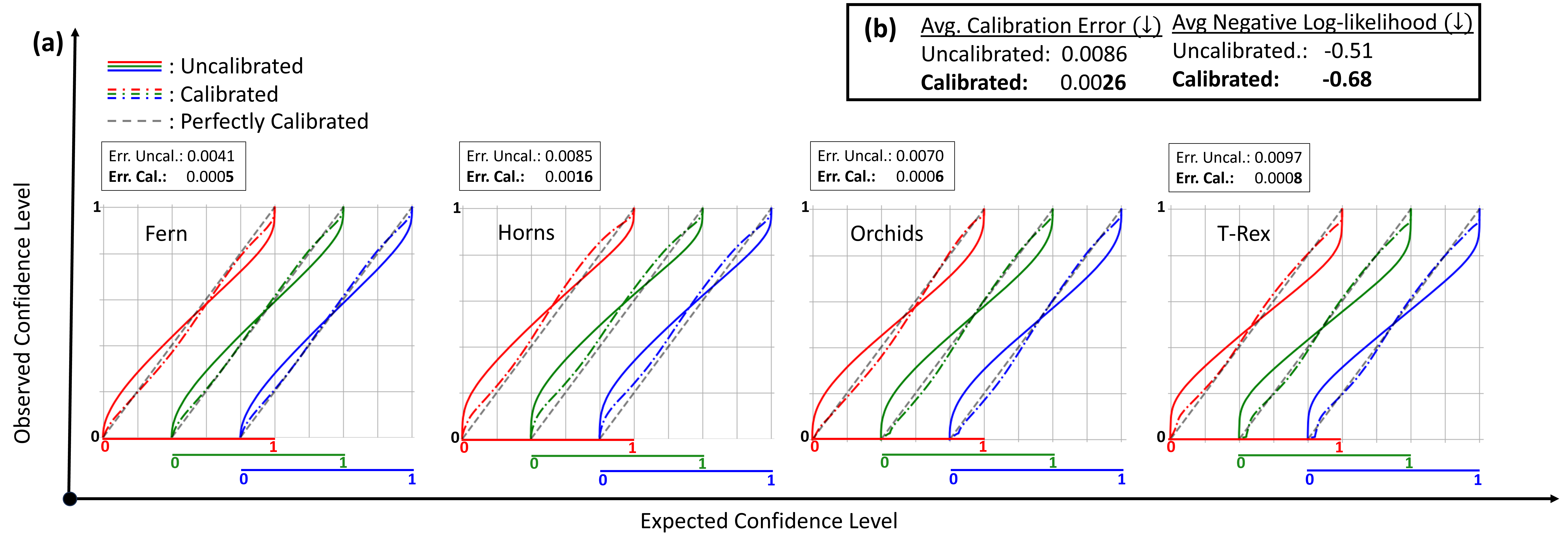}
    \vspace{-4mm}
    \caption{\label{fig:rgb_cal}\textbf{Quantitative comparison of uncalibrated and calibrated uncertainties.} In (a), we show calibration curves on test data from four scenes in LLFF \cite{llff}. The color of each curve indicates the color channel it corresponds to. The calibrated curves are much closer to the ideal calibration (dashed lines), demonstrating that the meta-calibrator works very well. In (b), the average calibration error and negative log-likelihood before and after calibration are reported for LLFF, clearly showing the meta-calibrator improves the accuracy of the uncertainties (lowering calibration error and negative log-likelihood). To test generalization, the meta-calibrator was also applied to held-out scenes in DTU \cite{dtu}, achieving a 70 \% reduction in calibration error on average.}
\end{figure*}

\begin{table*}
    \begin{center}
    \scriptsize
    \caption{\label{table:uncert_sota} \textbf{Quantitative results on standard NeRF uncertainty estimation benchmark.} Here, we present results on the LLFF \cite{llff} dataset used in prior work \cite{cf-nerf, s-nerf, da-nerf, wildnerf} on uncertainty estimation for NeRFs. This dataset is less challenging than the one in \cref{table:sota} since 4-12 views are used for training instead of 3. Our proposed approach results in significantly better uncertainties than prior methods for NeRF uncertainty estimation on all 8 scenes in LLFF. Note: \textbf{\emph{lower negative log-likelihood values indicate more accurate uncertainties.}} $M$ indicates the number of ensemble members, and MC-DO refers to Monte Carlo Dropout sampling with $M$ sample configurations. This table is Tab. 1 from \cite{da-nerf} with our results added as an additional column. Please refer to \cite{da-nerf} for further details.}
    \vspace{-3mm}
    \begin{NiceTabular}{c|c|c|c|c|c|c|c|c|c} 
     \hline
     \multicolumn{10}{c}{Negative Log-likelihood ($\downarrow$)}\\
     Scene & \# of Train. & MC-DO & Naïve Ens. & NeRF-W  & S-NeRF  & CF-NeRF  & DANE \cite{da-nerf} & DANE \cite{da-nerf} & \textbf{Ours}\\
     & Views & $M = 5$ & $M = 5$ & \cite{wildnerf} & \cite{s-nerf} & \cite{cf-nerf} & $M = 5$ & $M = 10$ & \\
     \hline
     Fern & 4 & 4.90 & 2.47 & 2.16 & 2.01 & --- & -0.98 & -1.00 & \textbf{-1.41}\\
     Orchids & 5 & 5.74 & 2.23 & 2.24 & 1.95 & --- & -0.28 & -0.31 & \textbf{-0.84}\\
      Leaves & 5 & 2.72 & 2.66 & 0.79 & 0.68 & --- & 0.97 & 0.73 & \textbf{-1.19}\\
     Flower & 7 & 4.63 & 1.63 & 1.71 & 1.27 & --- & 1.00 & 0.85 & \textbf{-2.05}\\
     Fortress & 8 & 5.19 & 2.29 & 1.04 & -0.03 & --- & -1.30 & -1.30 & \textbf{-1.99}\\
     Room & 8 & 5.06 & 2.13 & 4.93 & 2.35 & --- & -1.35 & -1.35 & \textbf{-2.17}\\
     T-Rex & 11 & 4.10 & 2.28 & 1.91 & 1.37 & --- & -0.31 & -0.69 & \textbf{-1.49}\\
     Horns & 12 & 4.18 & 2.17 & 0.78 & 0.60 & --- & -0.55 & -0.66 & \textbf{-2.18}\\
     \hline
     \multicolumn{2}{c}{Avg.} & 4.57 & 2.23 & 1.95 & 1.27 & 0.57 & -0.35 & -0.47 & \textbf{-1.67}\\
    \hline
    \end{NiceTabular}
    \end{center}
    \vspace{-3mm}
\end{table*}

\begin{figure}
    \centering
    \begin{subfigure}{0.45\linewidth}
      \includegraphics[width=\linewidth]{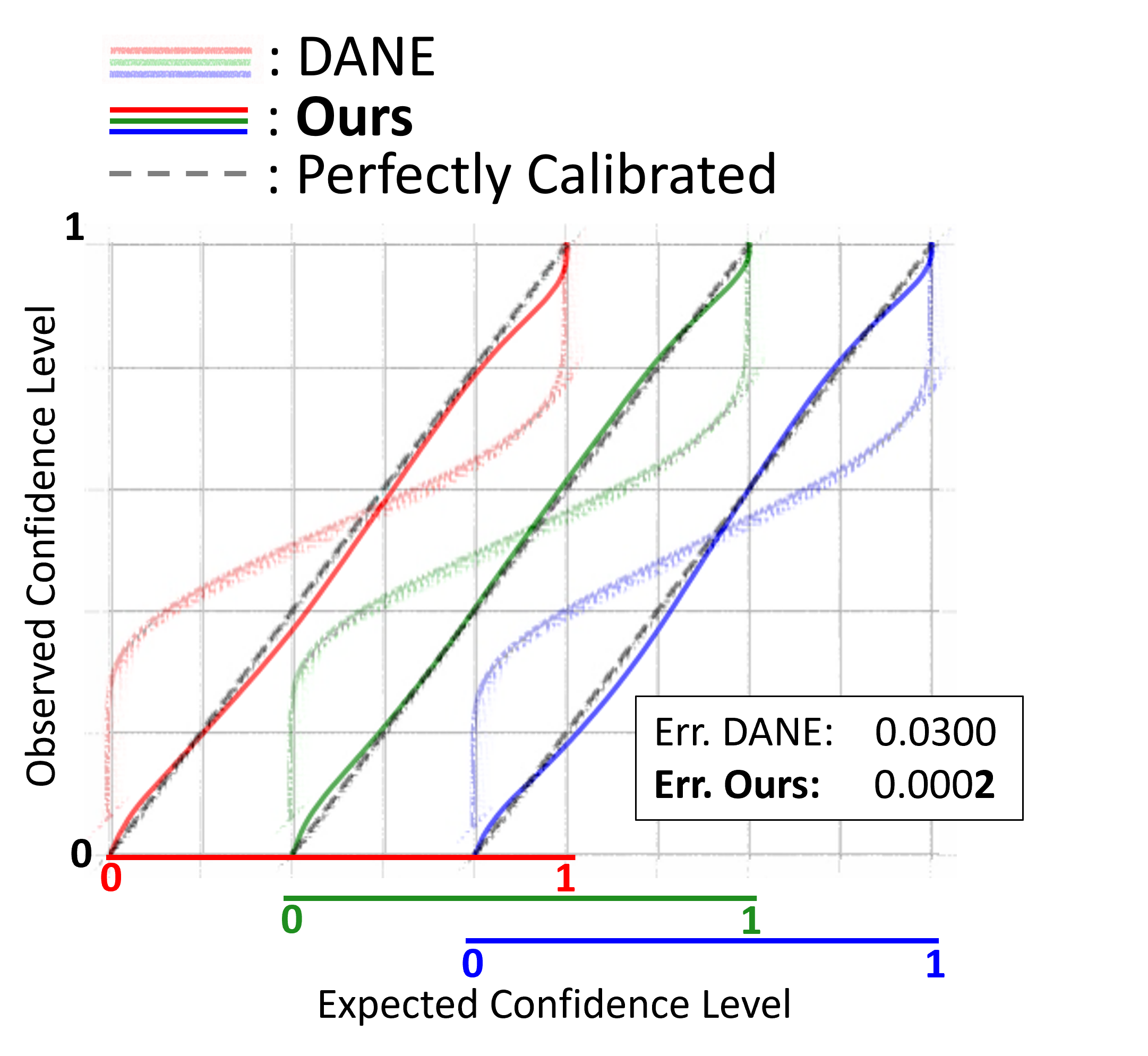}
      \caption{Calibration curves for the \emph{Fern} scene from 3-view LLFF \cite{llff}.}
      \label{fig:dane_curve_compare}
    \end{subfigure}
    \hfill
    \begin{subfigure}{0.45\linewidth}
      \includegraphics[width=\linewidth]{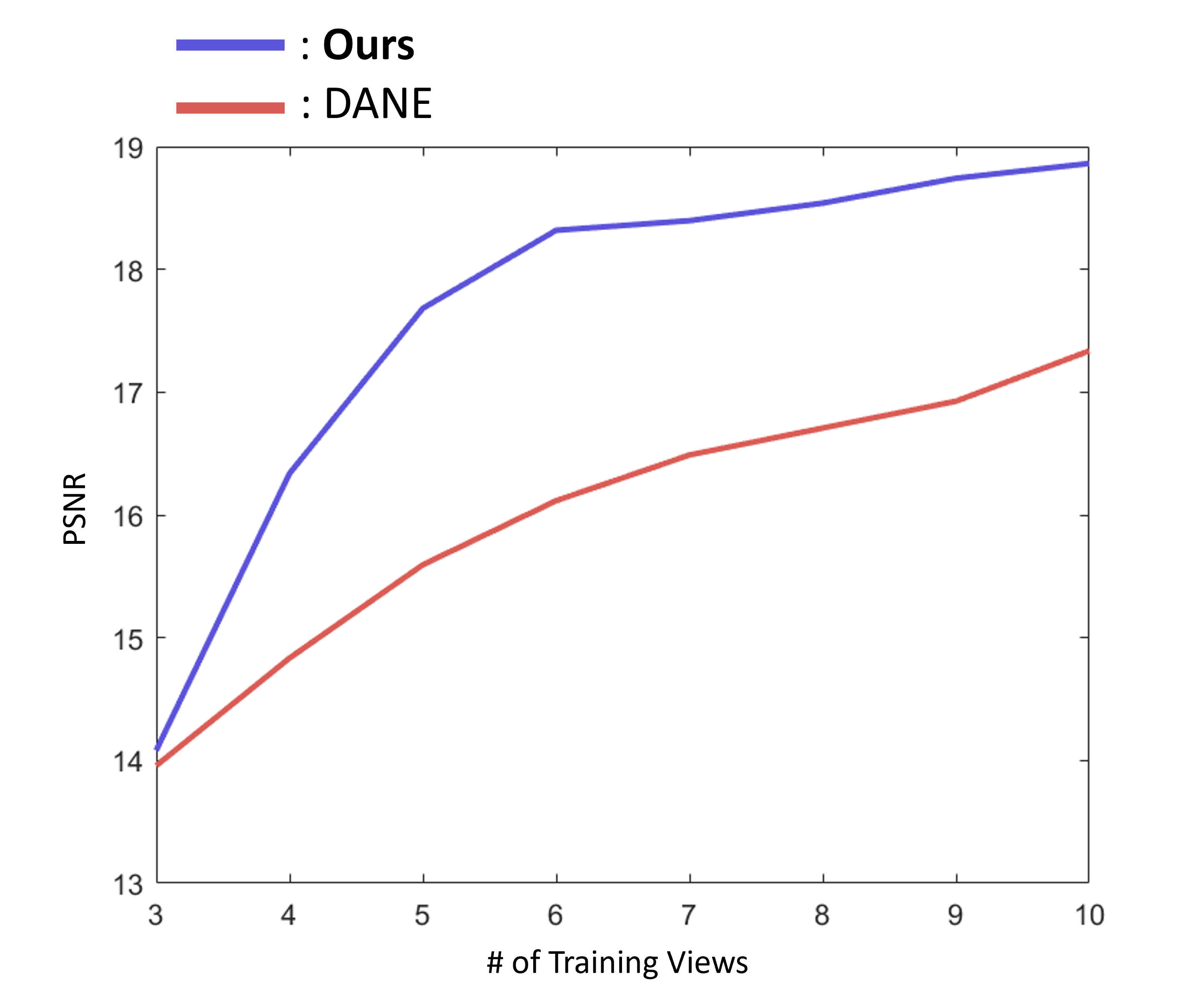}
      \caption{Next-best View Planning on \emph{Horns} from LLFF \cite{llff}.}
      \label{fig:nbv_dane}
    \end{subfigure}
    \caption{\textbf{Comparison to DANE \cite{da-nerf}.} Results comparing our uncertainties to those from the state-of-the-art method DANE. In (a) we show DANE's RGB calibration curves are not closely aligned with the perfectly calibrated lines, meaning it is miscalibrated. It is significantly over-confident for expected confidence levels close to 1 and under-confident for confidence levels close to 0. In comparison, the curves for our approach are extremely close to the ideal calibration (dashed lines), demonstrating that the meta-calibrator works very well, predicting expected confidences that match the true ones. This is also verified by how our calibration error is over two orders of magnitude smaller than DANE's. In (b) we show that our approach results in more efficient performance gains over DANE for next-best view planning.}
    \label{fig:dane_compare}
\end{figure}

\subsection{Comparison to Uncalibrated Uncertainties}
\label{sec:cal_compare}
\begin{figure*}
    \centering
    \includegraphics[width=\linewidth]{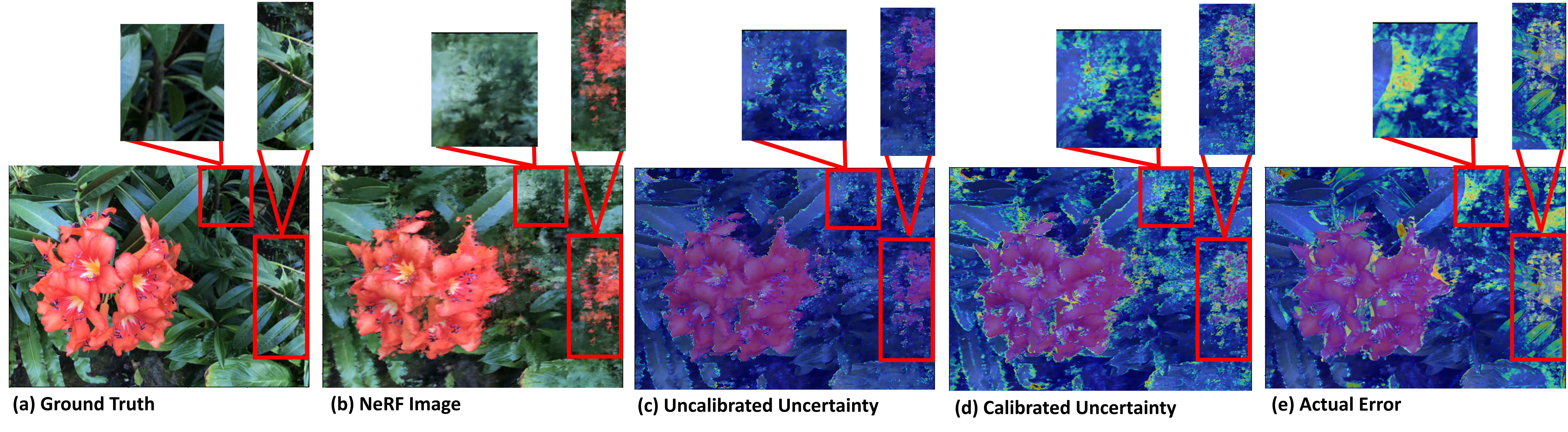}
    \caption{\label{fig:vis_uncert}\textbf{Qualitative comparison of uncalibrated and calibrated uncertainties from the \emph{Flower} scene in LLFF \cite{llff}.} The calibrated uncertainties (d) clearly detect incorrect regions (indicated by the red boxes) better than the uncalibrated uncertainties (c) do. This is apparent by noting that (d)  and (e) look more similar than (c) and (e).}
\end{figure*}

In this section, we compare our uncalibrated base NeRF uncertainties to our calibrated uncertainties obtained from applying the proposed meta-calibrator. In \cref{fig:vis_uncert} we show that the calibrated uncertainties better highlight floaters and other errors in the NeRF renderings. In \cref{fig:rgb_cal}, we show that the meta-calibrator predicts expected confidences that closely match the true ones, lowering both the calibration error and the negative log-likelihood of the uncalibrated uncertainties.

\subsection{Application: Next-best View Planning}
\label{sec:nbv_exp}
\begin{figure}[h!]
    \centering
    \includegraphics[width=\linewidth]{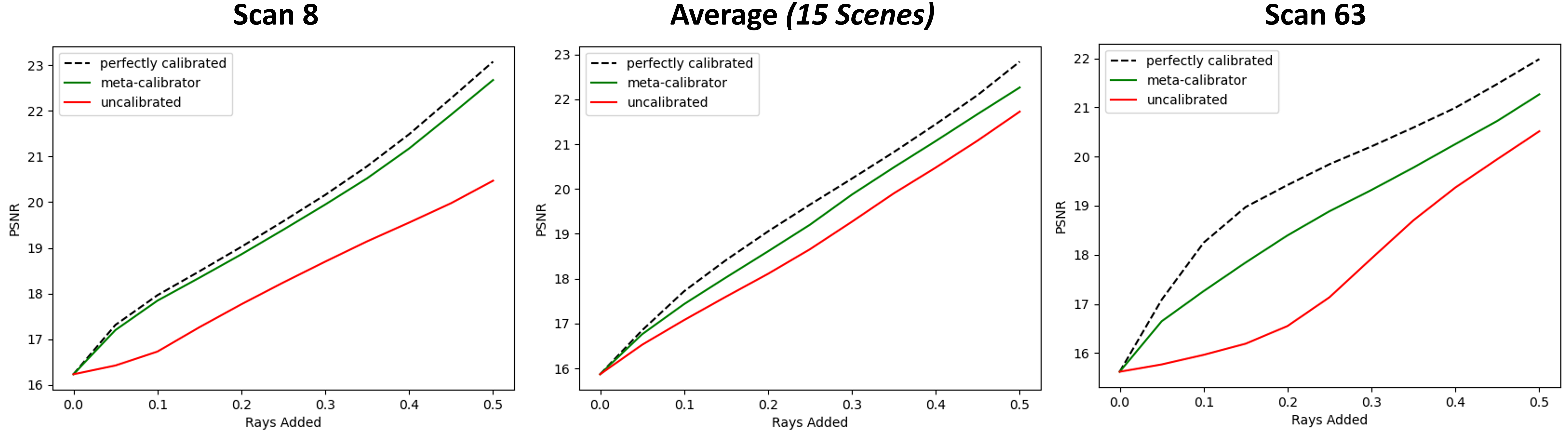}
    \caption{\textbf{Advantage of the meta-calibrator for next-best view planning}. This figure shows the information gain in DTU \cite{dtu} from uncalibrated and calibrated uncertainty-guided ray selection for next-best view planning. Picking rays according to the calibrated uncertainties (green) consistently results in higher PSNRs than picking rays according to the uncalibrated uncertainties (red). Individual results for \emph{Scan 8} (leftmost plot) and \emph{Scan 63} (rightmost plot) and average results over all fifteen scenes (middle plot) in DTU are shown. The dashed black lines show results for theoretically perfect calibration, where the ground truth calibration curves for the test set are used to construct the calibration curves instead of the meta-calibrator.}
    \label{fig:nbv_dtu}
\end{figure}
In this section, we show that our uncertainties can be leveraged for next-best view planning. Specifically, we start by training the NeRF model for 2000 iterations on a training set of three images. The next-best view is selected by evaluating the average calibrated pixel uncertainty (obtained using the meta-calibrator) for each of the candidate views, and the view with the highest uncertainty is added to the training set. The average PSNR of the test images is reported after each training iteration. In \cref{fig:nbv_dane} we show that using our approach results in greater performance gains (higher PSNRs) than DANE \cite{da-nerf} does.\\

To show that calibration, specifically, helps an agent select views that have the most potential for improving the NeRF's performance, we compare the information gain from rays selected according to the highest calibrated uncertainties to the information gain from rays selected according to the highest uncalibrated uncertainties. Specifically, for evenly spaced fractions $\gamma_{i} \in [0, 0.5]$, we plot the average PSNR over the test set assuming the top $100 \% \times \gamma_{i}$ most uncertain pixel colors are predicted perfectly by the NeRF model. We use the updated PSNR to quantify information gain. Intuitively, better uncertainties should result in selecting pixels with higher information gain. We show that the uncertainties calibrated by our meta-calibrator produce higher average PSNRs on the test set for scenes in DTU \cite{dtu} than the uncalibrated uncertainties in \cref{fig:nbv_dtu}. This shows that calibration specifically re-orders the pixel uncertainties so that rays more likely to raise the PSNR are picked earlier in next-best view planning. In \cref{sec:order}, we include a detailed theoretical example showing that such re-ordering is possible with our NeRF calibration procedure.

\section{Conclusion}
\label{sec:conclusion}
In this paper we addressed the open problem of obtaining calibrated uncertainties from NeRF models. We introduce the concept of a meta-calibrator that infers the calibration curves from scene features, and using this approach achieve state-of-the-art uncertainty without holding out any ground truth data from the target scene. By enabling efficient and accurate calibration of NeRF models without relying on additional data, our method represents a significant step forward in the practical application of NeRF to real-world scenarios and opens up new avenues for the use of NeRF in situations where data is limited and uncertainty is critical.

\paragraph{\textbf{Acknowledgements.}}

The authors would like to thank Seunghyeon Seo for his support on FlipNeRF, and Oishi Deb for her insightful feedback on uncertainty estimation. Niki Amini-Naieni is funded by an AWS Studentship, the Reuben Foundation, and the AIMS CDT program at the University of Oxford. Tomas Jakab is sponsored by EPSRC VisualAI EP/T028572/1 and Andrea Vedaldi by ERC-CoG UNION 101001212.

\paragraph{\textbf{Ethics.}}

We used the Realistic Synthetic 360$^{\circ}$ dataset \cite{nerf}, the subset of scenes in DTU \cite{dtu} used in \cite{flipnerf}, and all the scenes in LLFF \cite{llff} following their terms and conditions. There is no personal data. For further details on ethics, data protection, and copyright please see \url{https://www.robots.ox.ac.uk/~vedaldi/research/union/ethics.html}.

%
%
\bibliographystyle{splncs04}
\bibliography{main}

\begin{thebibliography}{10}
\providecommand{\url}[1]{\texttt{#1}}
\providecommand{\urlprefix}{URL }
\providecommand{\doi}[1]{https://doi.org/#1}

\bibitem{mipnerf}
Barron, J.T., Mildenhall, B., Tancik, M., Hedman, P., Martin-Brualla, R., Srinivasan, P.P.: Mip-nerf: A multiscale representation for anti-aliasing neural radiance fields. In: ICCV. pp. 5835--5844 (2021)

\bibitem{bhalgat2022hashnerfpytorch}
Bhalgat, Y.: Hashnerf-pytorch. \url{https://github.com/yashbhalgat/HashNeRF-pytorch/} (2022)

\bibitem{dynanerf}
Gafni, G., Thies, J., Zollh{\"o}fer, M., Nie{\ss}ner, M.: Dynamic neural radiance fields for monocular 4d facial avatar reconstruction. In: CVPR. pp. 8649--8658 (2021)

\bibitem{caluncdeep}
Ghoshal, B., Tucker, A.: On calibrated model uncertainty in deep learning. In: ECML (2022)

\bibitem{bayesrays}
Goli, L., Reading, C., Sellán, S., Jacobson, A., Tagliasacchi, A.: {Bayes' Rays}: Uncertainty quantification in neural radiance fields. ArXiv  \textbf{abs/2309.03185} (2023)

\bibitem{dietnerf}
Jain, A., Tancik, M., Abbeel, P.: Putting nerf on a diet: Semantically consistent few-shot view synthesis. In: ICCV. pp. 5865--5874 (2021)

\bibitem{nerf_mri}
Jang, T.J., Hyun, C.M.: Nerf solves undersampled mri reconstruction. ArXiv  \textbf{abs/2402.13226} (2024)

\bibitem{dtu}
Jensen, R.R., Dahl, A., Vogiatzis, G., Tola, E., Aan{\ae}s, H.: Large scale multi-view stereopsis evaluation. In: CVPR (2014)

\bibitem{nextbestview}
Jin, L., Chen, X., Ruckin, J., Popovi'c, M.: Neu-nbv: Next best view planning using uncertainty estimation in image-based neural rendering. In: IROS (2023)

\bibitem{volrend}
Kajiya, J.T., Von~Herzen, B.P.: Ray tracing volume densities. In: SIGGRAPH. p. 165–174 (1984)

\bibitem{nerfvae}
Kosiorek, A.R., Strathmann, H., Zoran, D., Moreno, P., Schneider, R., Mokr'a, S., Rezende, D.J.: Nerf-vae: A geometry aware 3d scene generative model. In: ICML (2021)

\bibitem{CalRegr}
Kuleshov, V., Fenner, N., Ermon, S.: Accurate uncertainties for deep learning using calibrated regression. In: ICML. pp. 2796--2804 (2018)

\bibitem{sparsevoxelnerf}
Liu, L., Gu, J., Lin, K.Z., Chua, T.S., Theobalt, C.: Neural sparse voxel fields. In: NIPS (2020)

\bibitem{wildnerf}
Martin-Brualla, R., Radwan, N., Sajjadi, M.S.M., Barron, J.T., Dosovitskiy, A., Duckworth, D.: Nerf in the wild: Neural radiance fields for unconstrained photo collections. In: CVPR (2021)

\bibitem{llff}
Mildenhall, B., Srinivasan, P.P., Ortiz-Cayon, R., Kalantari, N.K., Ramamoorthi, R., Ng, R., Kar, A.: Local light field fusion: Practical view synthesis with prescriptive sampling guidelines. In: TOG (2019)

\bibitem{nerf}
Mildenhall, B., Srinivasan, P.P., Tancik, M., Barron, J.T., Ramamoorthi, R., Ng, R.: Nerf: Representing scenes as neural radiance fields for view synthesis. In: ECCV (2020)

\bibitem{instant_ngp}
M\"uller, T., Evans, A., Schied, C., Keller, A.: Instant neural graphics primitives with a multiresolution hash encoding. In: ACM Trans. Graph. (2022)

\bibitem{donerf}
Neff, T., Stadlbauer, P., Parger, M., Kurz, A., Mueller, J.H., Chaitanya, C.R.A., Kaplanyan, A., Steinberger, M.: Donerf: Towards real-time rendering of compact neural radiance fields using depth oracle networks. In: CGF. pp. 45--59 (2021)

\bibitem{regnerf}
Niemeyer, M., Barron, J.T., Mildenhall, B., Sajjadi, M.S.M., Geiger, A., Radwan, N.: Regnerf: Regularizing neural radiance fields for view synthesis from sparse inputs. In: CVPR (2022)

\bibitem{oquab2023dinov2}
Oquab, M., Darcet, T., Moutakanni, T., Vo, H.V., Szafraniec, M., Khalidov, V., Fernandez, P., Haziza, D., Massa, F., El-Nouby, A., Howes, R., Huang, P.Y., Xu, H., Sharma, V., Li, S.W., Galuba, W., Rabbat, M., Assran, M., Ballas, N., Synnaeve, G., Misra, I., Jegou, H., Mairal, J., Labatut, P., Joulin, A., Bojanowski, P.: Dinov2: Learning robust visual features without supervision (2023)

\bibitem{animanerf}
Peng, S., Dong, J., Wang, Q., Zhang, S., Shuai, Q., Zhou, X., Bao, H.: Animatable neural radiance fields for modeling dynamic human bodies. In: ICCV (2021)

\bibitem{neurar}
Ran, Y., Zeng, J., He, S., Li, L., Chen, Y., Lee, G.H., Chen, J., Ye, Q.: Neurar: Neural uncertainty for autonomous 3d reconstruction. In: RAL (2023)

\bibitem{derf}
Rebain, D., Jiang, W., Yazdani, S., Li, K., Yi, K.M., Tagliasacchi, A.: Derf: Decomposed radiance fields. In: CVPR. pp. 14148--14156 (2020)

\bibitem{flipnerf}
Seo, S., Chang, Y., Kwak, N.: Flipnerf: Flipped reflection rays for few-shot novel view synthesis. In: ICCV (2023)

\bibitem{mixnerf}
Seo, S., Han, D., Chang, Y., Kwak, N.: Mixnerf: Modeling a ray with mixture density for novel view synthesis from sparse inputs. In: CVPR. pp. 20659--20668 (2023)

\bibitem{cf-nerf}
Shen, J., Agudo, A., Moreno-Noguer, F., Ruiz, A.: Conditional-flow nerf: Accurate 3d modelling with reliable uncertainty quantification. In: ECCV (2022)

\bibitem{s-nerf}
Shen, J., Ruiz, A., Agudo, A., Moreno-Noguer, F.: Stochastic neural radiance fields: Quantifying uncertainty in implicit 3d representations. In: 3DV. pp. 972--981 (2021)

\bibitem{nerfperact}
Sitzmann, V., Martel, J.N., Bergman, A.W., Lindell, D.B., Wetzstein, G.: Implicit neural representations with periodic activation functions. In: NIPS (2020)

\bibitem{da-nerf}
Sünderhauf, N., Abou-Chakra, J., Miller, D.: Density-aware nerf ensembles: Quantifying predictive uncertainty in neural radiance fields. In: ICRA (2023)

\bibitem{f2nerf}
Wang, P., Liu, Y., Chen, Z., Liu, L., Liu, Z., Komura, T., Theobalt, C., Wang, W.: F2-nerf: Fast neural radiance field training with free camera trajectories. In: CVPR (2023)

\bibitem{nerfplusplus}
Zhang, K., Riegler, G., Snavely, N., Koltun, V.: Nerf++: Analyzing and improving neural radiance fields. ArXiv  \textbf{abs/2010.07492} (2020)

\bibitem{lpips}
Zhang, R., Isola, P., Efros, A.A., Shechtman, E., Wang, O.: The unreasonable effectiveness of deep features as a perceptual metric. In: CVPR (2018)

\end{thebibliography}
\newpage
\appendix

In the supplementary material for \emph{Instant Uncertainty Calibration of NeRFs Using a Meta-calibrator}, we include additional details on the motivation for the meta-calibrator, applications of our approach, and experiments and code to support our design. In \cref{sec:pca_nec} we explain why the Principal Component Analysis (PCA) representation of the calibration curves is necessary; in \cref{sec:train_curve_overfit} we show that using the training set as the calibration set results in severe overfitting; in \cref{sec:hold_out_data_poor}, we show that holding out data results in poor performance at image reconstruction; in \cref{sec:num_samples}, we investigate the influence of the number of samples along the ray on the quality of the base uncertainties; in \cref{sec:order} we include a detailed example showing that calibration can re-order the pixel uncertainties, improving applications such as next-best view planning (see \cref{fig:nbv_dtu}); and in \cref{sec:efficiency} we demonstrate the efficiency of our uncertainty metric over other approaches. The additional details, explanations, experiments, and code provided here are intended to enhance the reader's understanding of our approach and to further motivate, support, and explain the statements in the main paper. In summary, the supplementary material complements the content in the main paper and answers potential lingering questions such as why a low-dimensional representation of the calibration curves was chosen over a high-dimensional one.

\section{Why is the PCA Representation Necessary?}
\label{sec:pca_nec}
One might wonder why the PCA parameterization of the curves is necessary - why not simply predict a discretized representation of the curve directly? In essence, the PCA parameterization of the calibration curves allows us to simplify the complex, high-dimensional data into a low-dimensional, manageable form.  This approach is favored over direct prediction of the calibration curve primarily because it is difficult to predict a high-dimensional output without a large amount of training data. The low-dimensional representation therefore improves the model's generalization capabilities for new scenes. Even in cases where a large amount of training data might be available, it is unnecessary to learn this from data because, as we show in Figure 3a in the main paper and Figure \ref{fig:curves_example} in the appendix, the calibration curves themselves lie on a low-dimensional subspace. To further motivate the use of the PCA, we show an example where we compare predicting the PCA coefficients to directly predicting a discretized 384-dim representation of the curve (with an MLP of size [128,128,384]). From \cref{fig:pca_vs_direct_pred} we see that the direct prediction (red "Discretized" curve ) leads to a noisier curve with higher error.

\begin{figure*}[h!]
    \centering
    \includegraphics[width=1\linewidth]{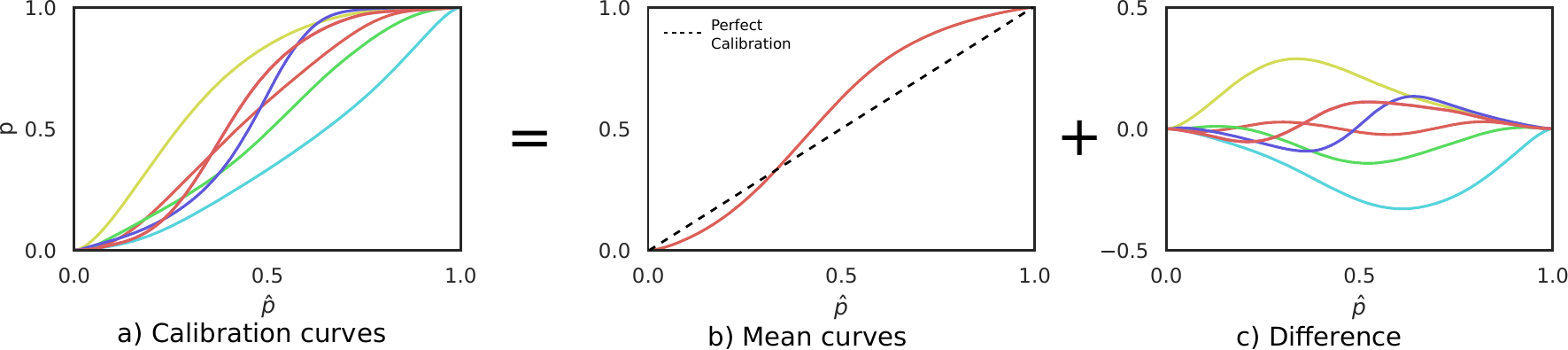}
    \caption{\textbf{Regularity in the calibration curves:} This figure shows the calibration curves obtained for seven of the real-world \textit{DTU dataset} scenes \cite{dtu}. While the calibration curves vary significantly across scenes there is a high degree of regularity in this variation. We use this insight to construct a low-dimensional parameterization of the curves that our meta-calibrator can predict from scene features.}
    \label{fig:curves_example}
\end{figure*}

\begin{figure}[h!]
    \centering
    \includegraphics[width=0.5\linewidth]{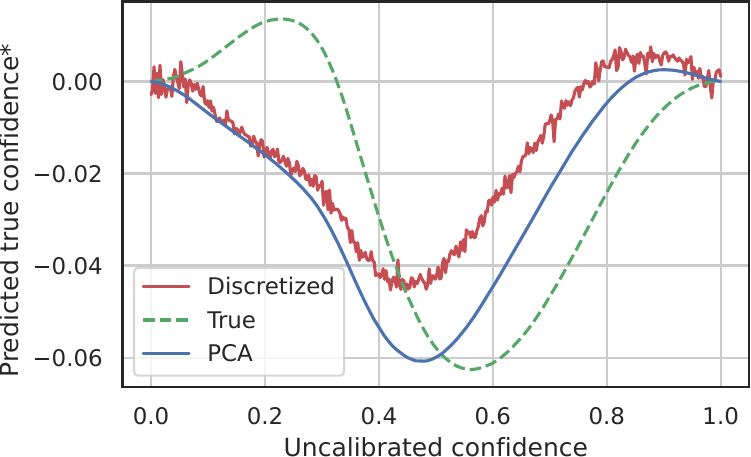}
    \caption{\textbf{PCA representation vs direct prediction of the calibration curve.} Here we show an example \textit{mean-normalized} (*) calibration prediction for the \textit{Fern} scene from the LLFF~\cite{llff} dataset. The low-dimensional PCA parameterization (blue "PCA" curve) allows the model to generalize better and better preserves the characteristics of the true calibration curves than the high-dimensional representation (red, noisy "Discretized" curve) does.}
    \label{fig:pca_vs_direct_pred}
\end{figure}
\section{Using the Training Set Leads to Severe Overfitting}
\label{sec:train_curve_overfit}
\begin{figure}[h!]
    \centering
    \includegraphics[width=\linewidth]{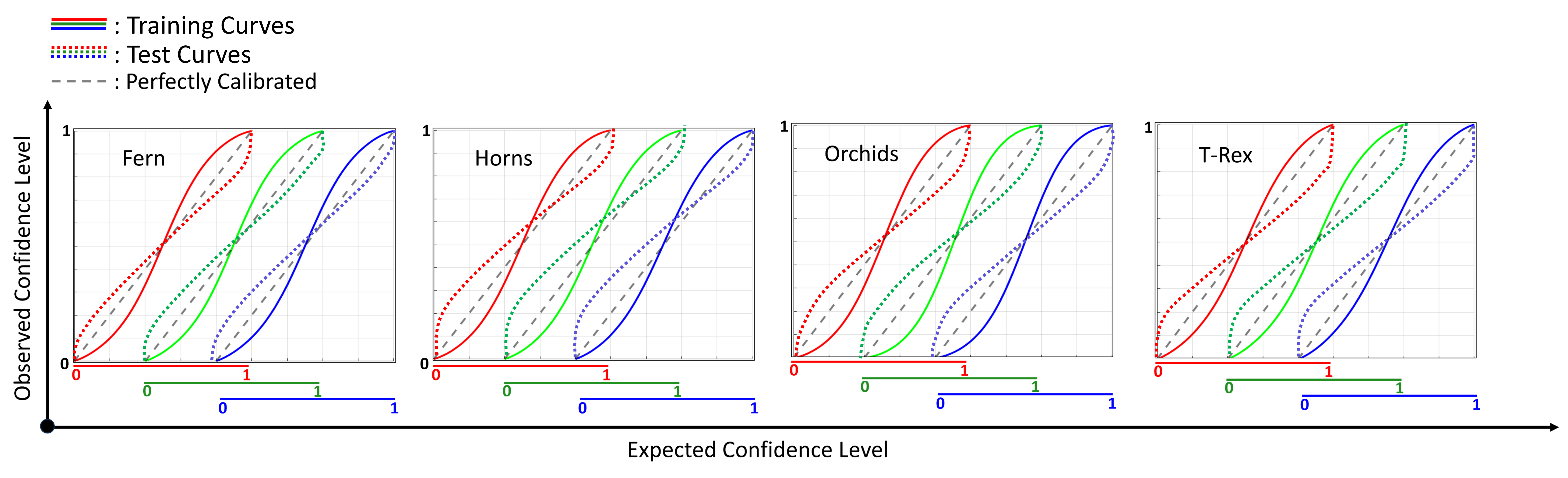}
    \caption{\textbf{Calibration curves for training and test data from four scenes in LLFF \cite{llff}}. The color of each curve indicates the color channel it corresponds to. The solid red, green, and blue curves are not closely aligned with the grey dashed lines in general, showing that the pretrained NeRF model is miscalibrated, even for the training set. The expected confidence levels for the training set (solid RGB lines) follow a different pattern from the one followed by the test set (dotted RGB lines). This is apparent by observing that the dotted RGB curves are not aligned with the solid RGB curves close to zero and one. As a result, calibration with the training set (solid curves) would not generalize to the test set (dotted curves).}
    \label{fig:training_curves}
\end{figure}

One might consider applying the calibration technique for regression in \cite{CalRegr} directly to NeRFs by fitting a new calibrator on the training set for each new scene instead of using the meta-calibrator introduced in our work. To show why this will not work, in this section, we present the calibration curves of the training rays for four scenes in LLFF \cite{llff} as the solid RGB curves in \cref{fig:training_curves}. These curves reveal that not only are the confidence levels of the pretrained NeRF model miscalibrated for the training set, but the pattern they follow is different from the one observed in the test set, also shown in \cref{fig:training_curves}. The NeRF model is consistently overconfident for confidence levels closer to zero and underconfident for confidence levels closer to one for the training set but consistently underconfident for confidence levels closer to zero and overconfident for confidence levels closer to one for the test set. Thus, calibration using the training set would result in very poor generalization to the test set. Specifically, using the training set for calibration results in worse test calibration errors than leaving the NeRF model uncalibrated for all scenes in LLFF.

\section{Holding Out Data Results in Poor Image Quality}
\label{sec:hold_out_data_poor}
While, as shown in \cref{sec:train_curve_overfit}, using the training set for calibration results in severe overfitting, we could also consider holding out images from the training set and using them to fit the calibrator. However, as shown in \cref{table:hold-out-data-poor-img-quality}, this method significantly reduces the performance of the NeRF at novel view synthesis. For example, holding out just one image from the \emph{Horns} scene in LLFF \cite{llff} reduces the PSNR by 17\%. Therefore, holding out images is not an ideal technique for fitting the calibrator. Unlike holding out images, our meta-calibrator allows the NeRF model to use all available data from the target scene for training, resulting in better image quality.\\

\begin{table}[h!]
\begin{center}
\scriptsize
\caption{\label{table:hold-out-data-poor-img-quality}\textbf{PSNR on 3-View LLFF \cite{llff} using 1, 2, and 3 views for training.} Holding out views from the training set significantly reduces quality of images inferred by NeRF. This is clear by observing how small PSNRs are in row 1 (training on 1 view) vs PSNRs in row 3 (training on 3 views). Higher PSNRs indicate better image quality. Note: NeRF model was trained for 2k iterations.}

\begin{NiceTabular}{c|c|c|c|c|c|c|c|c|c} 
 \hline
 Num. of Views & Room & Fern & Flower & Fortress & Horns & Leaves & Orchids & T-Rex\\
\hline
1 & 15.94 & 16.15 & 12.93 & 15.19 & 12.58 & 11.87 & 10.99 & 11.24 \\
2 & 18.57 & 19.07 & 17.38 & 17.45 & 13.51 & 14.12 & 14.37 & 17.36 \\
3 & \textbf{19.25} & \textbf{19.76} & \textbf{18.06} & \textbf{21.12} & \textbf{16.28} & \textbf{15.44} & \textbf{15.72} & \textbf{18.33} \\
\hline
\end{NiceTabular}

\end{center}
\end{table}
\section{Number of Ray Samples' Influence on Uncertainty Quality}
\label{sec:num_samples}
\begin{figure}[h!]
    \centering
    \includegraphics[width=0.5\linewidth]{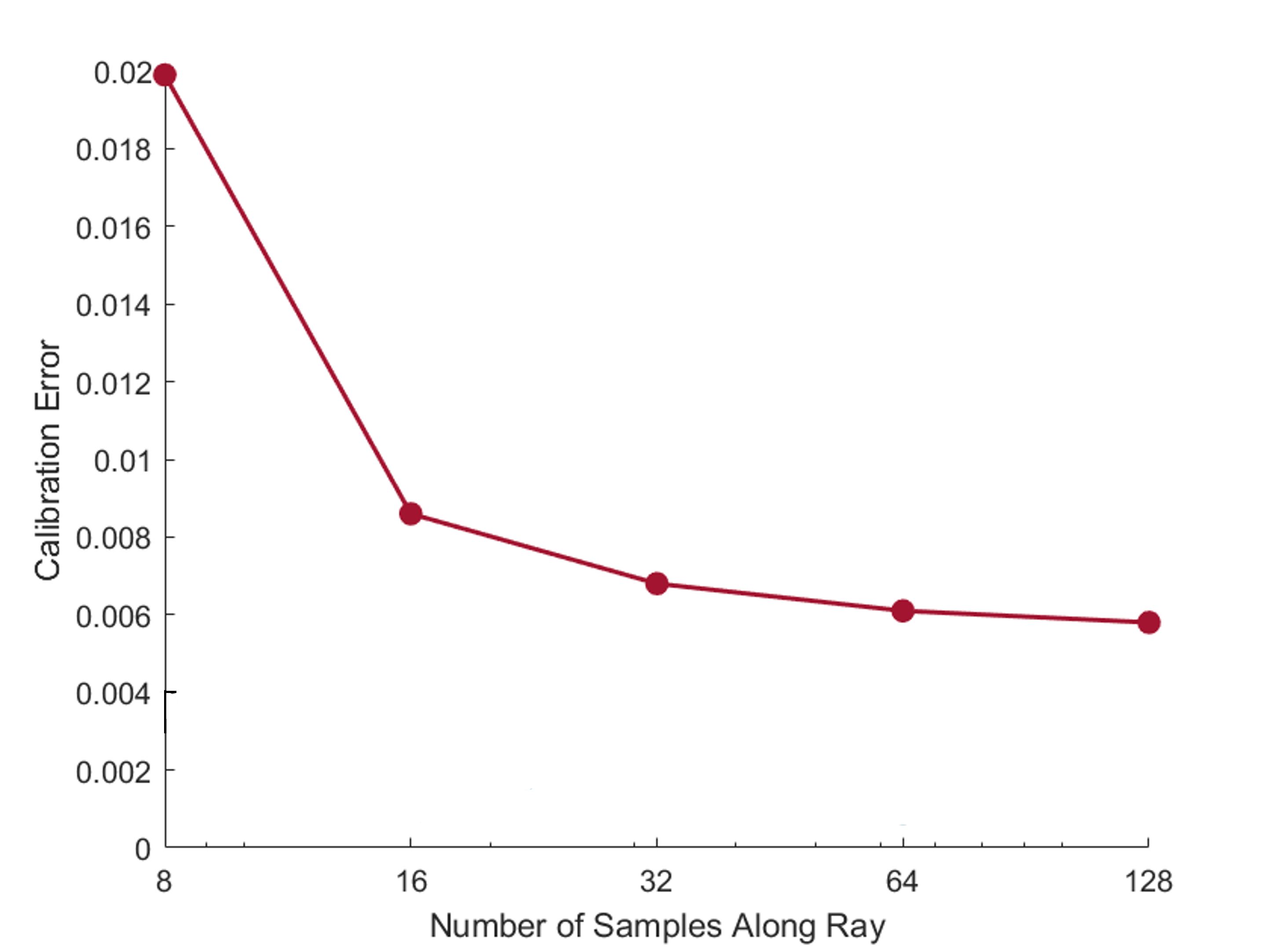}
    \caption{\textbf{Uncalibrated uncertainty quality vs number of samples along a ray for \emph{Room} scene from LLFF \cite{llff}.} The number of samples along the ray for FlipNeRF \cite{flipnerf} determines the number of components in the Laplacian mixture model used to obtain the uncertainty in the predicted images. Higher number of samples increases the precision of the mixture model, reducing the calibration error of the base uncertainties.}
    \label{fig:num_samples}
\end{figure}

The number of samples along the ray for FlipNeRF \cite{flipnerf} determines the number of components in the Laplacian mixture model used to represent the uncertainty in the predicted images. Intuitively, increasing the number of mixture components, and, hence, the number of ray samples, increases the precision of this representation. Supporting this concept, we show in \cref{fig:num_samples} that as the number of ray samples increases, the calibration error of the base uncertainties decreases, with diminished returns after 128 samples. We use 128 ray samples for our pretrained FlipNeRF model as this produces the lowest calibration error for the base uncertainties without being as costly to train as NeRFs with higher sample counts.
\section{Calibration Can Correct the Order of Pixel Uncertainties}
\label{sec:order}
Our meta-calibrator can re-order the pixel uncertainties even though it predicts a monotonic regression model that maps the NeRF's expected confidences to the true ones. Here, we include a detailed theoretical example showing that such re-ordering is possible.\\

\begin{figure*}[h!]
\centering
\includegraphics[width=\linewidth]{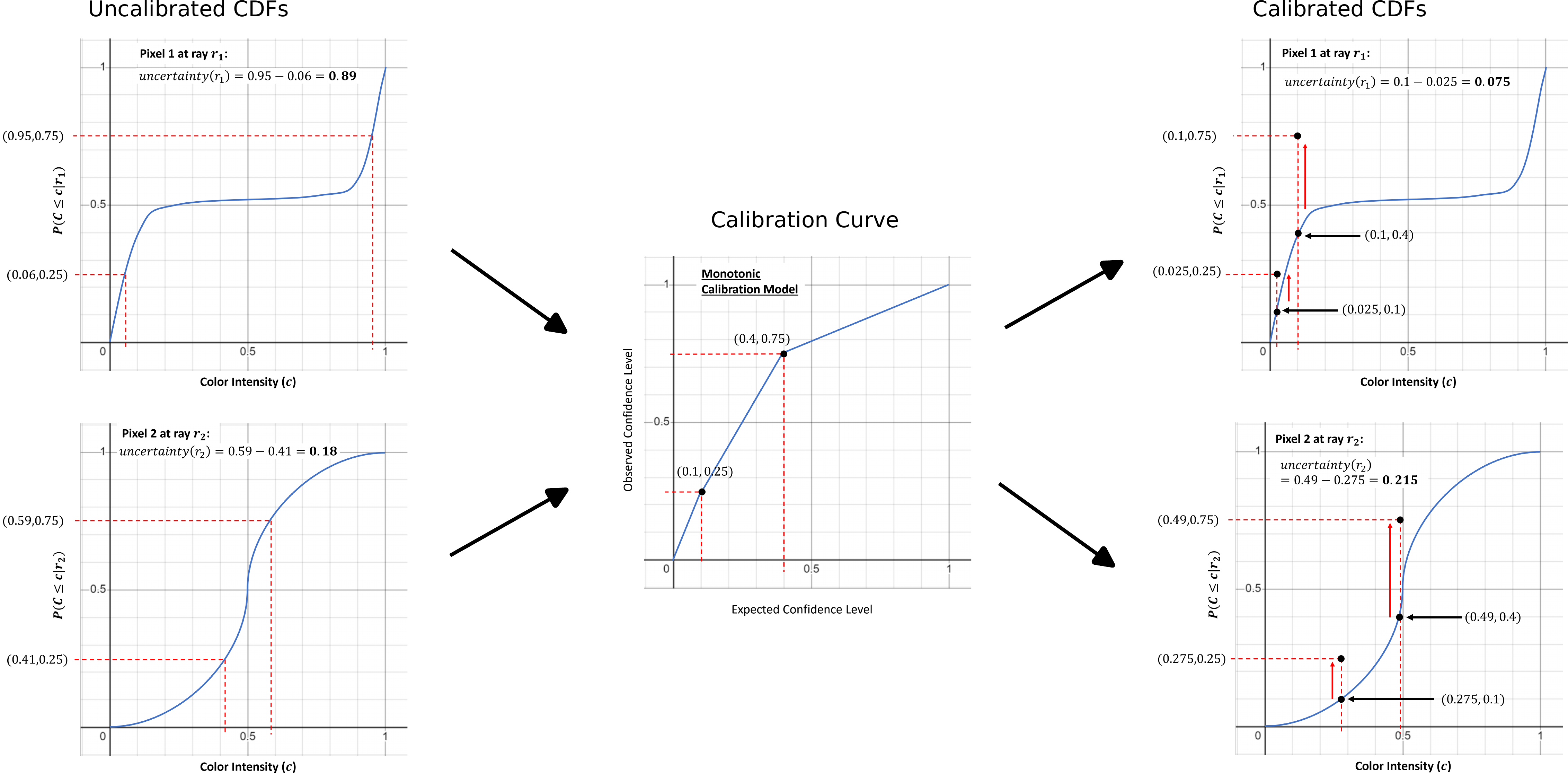}
\caption{\textbf{Is the order of uncertainties necessarily preserved during calibration?} This illustration shows that the order of uncertainties for two pixels (corresponding to rays 1 and 2) is not necessarily preserved during calibration. We start with $uncertainty(r_{1}) > uncertainty(r_{2})$ for the left two uncalibrated CDFs and end up with $uncertainty(r_{1}) < uncertainty(r_{2})$ after calibration on the right.}
\label{fig:order-explanation}
\end{figure*}

One might think that the order of the uncertainties is preserved by calibration as we're fitting a monotonic curve to the expected confidences. However, this is not the case. \textit{Rather than preserving the order of the uncertainties with respect to the pixels, the calibration preserves the monotonicity of the individual CDFs at each pixel.} To elucidate this concept, consider an example where calibration can reverse the order of uncertainties for two pixel CDFs as shown in Figure \ref{fig:order-explanation}.

Initially, the CDF for ray 1, corresponding to pixel 1, might indicate higher uncertainty compared to ray 2 (pixel 2). However, after applying the calibration process, the order of uncertainties can be reversed. This reversal is attributed to the differing shapes and slopes of the CDFs, which are altered non-linearly during calibration. The implications of this observation are significant. It underscores the non-trivial nature of the calibration process in uncertainty modeling and suggests that calibration does not merely scale or shift uncertainties but can fundamentally alter the relation between the uncertainty values. In summary, this highlights the complexity and nuanced impact of calibration on the predicted uncertainties.

\section{Efficiency of Uncertainty Metric}
\label{sec:efficiency}
In this section, we provide further details on why we use the interquartile range, rather than the variance or standard deviation, to quantify the uncertainty at each pixel. While the variance of a Laplacian mixture model can be obtained in closed form from the parameters of the component distributions, in our approach, the parameters of the \emph{calibrated} CDF (e.g., the location and scale parameters of the component CDFs) are not known. Hence, to obtain the variance of the distribution for each pixel, we would either need to sample from it or differentiate the predicted CDF to obtain the corresponding PDF and then integrate to estimate the variance. As shown in Table \ref{table:timing}, both of the aforementioned methods are much slower than estimating the interquartile range. This is because the interquartile range can be calculated from the calibrated CDF directly.

\begin{table}[h!]
\begin{center}
\scriptsize
\caption{\label{table:timing}\textbf{Timing of obtaining different uncertainty measures for the distribution of 1 pixel.} Calculating the interquartile range is much faster than calculating the variance.}
\begin{NiceTabular}{|c|c|c|} 
 \hline
 Uncertainty Metric & Method & Time (s)\\
  \hline
Variance & Integration & 9.807\\
Variance & Sampling & 1.759\\
\textbf{Interquartile Range (Ours)} & \textbf{Interpolation} & \textbf{0.008}\\
 \hline
\end{NiceTabular}
\end{center}
\vspace{-4mm}
\end{table}
\end{document}